\PassOptionsToPackage{numbers,sort&compress}{natbib}

\documentclass[a4paper,fleqn]{cas-dc}

\usepackage[numbers]{natbib}

\usepackage{stfloats} 
\usepackage{float}

\usepackage{hyperref} 
\usepackage[all]{hypcap} 

\makeatletter
\def\printorcid#1{}
\makeatother

\def\tsc#1{\csdef{#1}{\textsc{\lowercase{#1}}\xspace}}
\tsc{WGM}
\tsc{QE}
\tsc{EP}
\tsc{PMS}
\tsc{BEC}
\tsc{DE}

\begin{document}
\let\WriteBookmarks\relax
\shorttitle{Homogeneous Semantic Alignment and Hierarchical Expert Routing for Radiology Report Generation}
\shortauthors{E. Zhang et~al.}

\title [mode = title]{Homogeneous Semantic Alignment and Hierarchical Expert Routing for Radiology Report Generation}                      

\author[1]{Erjian Zhang}
\fnmark[1]

\credit{Conceptualization of this study, Methodology, Software}

\affiliation[1]{organization={School of Computer Science and Technology, Xinjiang University},
                city={Urumqi},
                postcode={830017},
                state={Xinjiang},
                country={China}}

\author[1]{Jiayuan Ma}
\fnmark[1]

\author[1,2]{Liejun Wang}
\cormark[1]
\ead{wljxju@xju.edu.cn}

\credit{Data curation, Writing - Original draft preparation}

\affiliation[2]{organization={Xinjiang Multimodal Intelligent Processing and Information Security Engineering Technology Research Center},
                city={Urumqi},
                postcode={830017},
                state={Xinjiang},
                country={China}}
                
\author[1]{Yikemaiti Sataer}

\author[1,3]{Xiaoming Tao}

\affiliation[3]{organization={Department of Electronic Engineering, Tsinghua University},
                city={Beijing},
                postcode={100084},
                country={China}}

\author[1,2]{Zhiqing Guo}
\cormark[1]
\ead{guozhiqing@xju.edu.cn}

\cortext[cor1]{Corresponding authors.}
\fntext[1]{These authors contributed equally to this work.}

\begin{abstract}
Radiology report generation (RRG) aims to convert medical images into diagnostic texts to assist in clinical decision-making and alleviate the workload of physicians. Although existing methods have made extensive progress in cross-modal interaction and the incorporation of external priors, the distribution shift of underlying representations and the undifferentiated rigid coupling of heterogeneous information cause weak visual abnormality cues to be easily diluted by massive text priors and generation inertia during decoding. To overcome this bottleneck, inspired by cognitive science, we propose a novel two-stage \textbf{H}omogeneous \textbf{S}emantic \textbf{A}lignment and \textbf{H}ierarchical \textbf{E}xpert \textbf{R}outing (HSA-HER) framework. First, the model introduces an explicit homogeneous distribution constraint in the underlying latent space to effectively eliminate the cross-modal distribution shift between visual and textual features, thereby extracting purified visual features as semantic anchors that accurately align with diseases. Second, for heterogeneous clinical evidence composed of visual features, local entities, and global retrievals, we design a hierarchical expert routing mechanism guided by these disease semantic anchors. This mechanism abandons the undifferentiated rigid coupling paradigm. Specifically, it dynamically activates expert networks to perform targeted mining and semantic reconstruction on multi-source evidence, and adaptively allocates fusion weights. Extensive experiments on three mainstream benchmark datasets demonstrate that HSA-HER achieves state-of-the-art performance, accurately depicting complex imaging details and key diagnostic information.
\end{abstract}



\begin{keywords}
Radiology report generation \sep Cross-modal alignment \sep Expert routing \sep Chest X-ray
\end{keywords}

\maketitle

\section{Introduction}

Radiology report generation (RRG) is a critical task in medical artificial intelligence. It aims to automatically convert medical images into diagnostic texts, addressing the growing demand for image interpretation and alleviating radiologists' workload. Efficient RRG frameworks optimize routine clinical workflows, accelerate patient screening, and assist in clinical decision-making.

To achieve this goal, existing studies primarily adopt the encoder-decoder paradigm. Early works focused on sequence fitting for long texts, but these methods are constrained by surface-level vision-text mappings and lack deep pathological semantic understanding \citep{chen2020generating, liu2021competence}. To address this, recent studies introduce large language models \citep{liu2024bootstrapping}, large-scale vision-language pre-training \citep{wang2025cxpmrg}, and retrieval-augmented mechanisms \citep{wang2026r2gencsr}, expanding feature representations and prior contexts. Although improving report quality, they still face bottlenecks in cross-modal feature alignment and heterogeneous information scheduling. First, existing cross-modal interactions mostly rely on attention fusion or instance-level contrastive learning \citep{wang2024camanet, cao2026tsmrg}. This shallow alignment fails to eliminate the inherent modality gap, leading to feature distribution shifts and hindering the extraction of pure visual features as disease-aligned semantic anchors. Second, incorporating multi-source medical priors and retrieval-augmented mechanisms \citep{wang2023metransformer, wu2026disease, wang2026activating} makes the input highly heterogeneous at the semantic granularity level. Without reliable semantic anchor guidance, existing methods typically adopt undifferentiated rigid coupling strategies, lack targeted reconstruction of abnormal semantics, and indiscriminately fuse these heterogeneous inputs into a single text generation path. Ultimately, weak visual abnormality cues are easily diluted during autoregressive decoding by massive text priors and generation inertia. This yields templated reports that struggle to accurately depict complex imaging details and key diagnostic information.


\begin{figure*}[t]
    \centering
    \includegraphics[width=\linewidth]{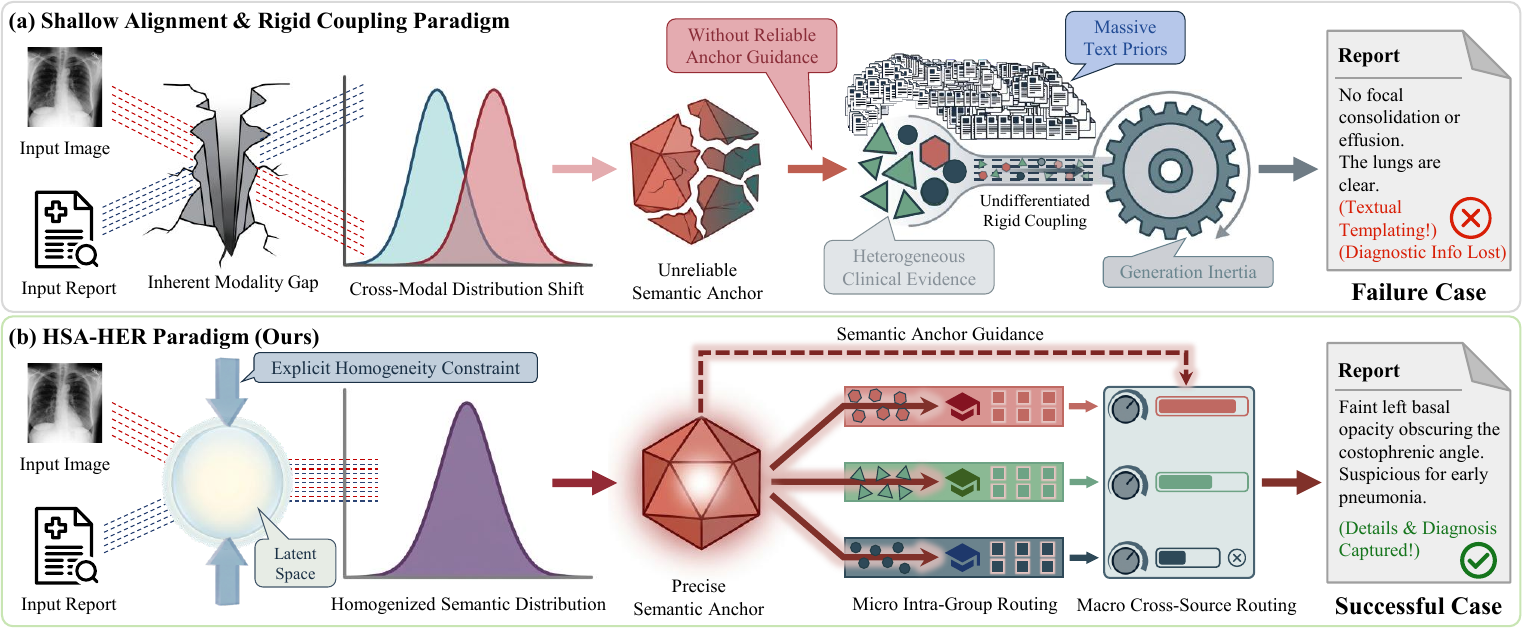}
    \caption{Comparison between the conventional paradigm and our HSA-HER framework. \textbf{(a)} The distribution shift of underlying representations and the undifferentiated rigid coupling of heterogeneous information cause weak visual abnormality cues to be easily diluted by massive text priors and generation inertia during decoding. \textbf{(b)} HSA-HER accurately depicts imaging details and diagnostic information through homogeneous semantic alignment and a hierarchical expert routing mechanism.}
    \label{fig:1}
\end{figure*}

Inspired by dual coding theory \citep{paivio1991dual} in cognitive science, we re-examine these limitations. This theory posits that visual and verbal cognition operate as independent representational systems, with high-level cognition relying on precise cross-system mappings between them. In clinical diagnosis, physicians bridge visual representations and textual knowledge to establish reliable disease anchors. Furthermore, to process complex cues, they avoid simple information stacking, instead invoking semantic qualification mechanisms to abstract and organize multidimensional cues into structured diagnostic evidence \citep{bordage1994elaborated}. This cognitive mechanism highlights two key dimensions to overcome existing RRG bottlenecks in feature alignment and information scheduling. Initially, to achieve substantial cross-system isomorphism, the model needs to introduce distribution constraints in the underlying latent space to eliminate cross-modal heterogeneity, thereby establishing disease-aligned semantic anchors. Subsequently, guided by these anchors, the model needs to discard existing undifferentiated rigid coupling methods to perform targeted semantic reconstruction on the heterogeneous clinical evidence (composed of visual features, local entities, and global retrievals) and reorganize the weights of different types of evidence.

Motivated by these observations, we propose a novel two-stage homogeneous semantic alignment and hierarchical expert routing framework, as illustrated in Figure \ref{fig:1}. In the first stage, we introduce an explicit homogeneous distribution constraint in the latent space. This forces heterogeneous visual and textual features into a strictly consistent distribution, eliminating cross-modal distribution shifts and extracting pure visual features as disease-aligned semantic anchors. In the second stage, we design a hierarchical expert routing mechanism to schedule multi-source information before decoding. This mechanism decouples the scheduling of this heterogeneous clinical evidence into micro and macro routing actions. At the micro intra-group routing level, guided by these disease semantic anchors, the model dynamically activates expert sub-networks for targeted mining and semantic reconstruction on each heterogeneous feature space, yielding highly specialized clinical representations. Subsequently, at the macro cross-source routing level, a top-level gating module uses these anchors to adaptively assign fusion weights to the purified multi-source evidence, discarding the undifferentiated rigid coupling paradigm. Ultimately, our framework prevents weak visual abnormality cues from being diluted by text priors and generation inertia, enabling the generated reports to accurately depict complex imaging details and key diagnostic information.

In summary, our main contributions are as follows:
\begin{itemize}
    \item We propose a novel RRG paradigm inspired by cognitive science, providing a fresh perspective of low-level representation homogeneity alignment and high-level structured evidence scheduling for this task.
    \item We introduce explicit homogeneous distribution constraints into the latent space. This eliminates cross-modal distribution shifts and extracts pure visual features to establish disease semantic anchors for subsequent multi-source information scheduling.
    \item We design a hierarchical expert routing mechanism guided by disease semantic anchors. It leverages micro intra-group routing to dynamically activate expert sub-networks for targeted mining and semantic reconstruction across heterogeneous feature spaces, and employs macro cross-source routing to adaptively allocate weights for multi-source clinical evidence.
    \item Our framework achieves superior performance across three mainstream benchmark datasets, successfully overcoming the bottleneck where weak visual abnormality cues are easily diluted by massive text priors and generation inertia.
\end{itemize}

\section{Related Work}

\subsection{Cross-modal feature alignment in RRG}

Radiology report generation (RRG) aims to translate medical images into accurate and coherent clinical texts. Unlike natural image captioning \citep{vinyals2015show}, RRG requires extracting weak pathological features, generating lengthy clinical reports, and maintaining strict medical logic. Early works primarily addressed these challenges by improving the long-sequence modeling capability for medical texts. For instance, R2Gen \citep{chen2020generating} and CMCL \citep{liu2021competence} utilized relational memory matrices and multimodal curriculum learning, respectively. However, these purely data-driven methods relied on surface-level image-text mappings, lacking an understanding of deep pathological semantics.

To deepen pathological semantic alignment, recent studies have explored various cross-modal interaction strategies. Early methods optimized image-text mappings via weakly supervised contrastive learning \citep{yan2021weakly} and reinforcement learning \citep{qin2022reinforced}. Subsequent works explicitly guided visual-textual alignment using class activation maps for attention supervision \citep{wang2024camanet} or concept-level matching via updatable medical knowledge bases \citep{yang2023radiology}. With the emergence of large language models (LLMs) and causal inference, cross-modal correspondence is now steered by diagnostic prompts \citep{jin2024promptmrg} or decoupled from confounding factors via causal interventions \citep{chen2025cross}. To further refine this alignment, recent models employ large-scale contrastive pre-training \citep{wang2025cxpmrg}, collaborative fusion networks for fine-grained lesion matching \citep{hou2026collaborative}, and hybrid contrastive strategies to capture temporal semantics in longitudinal data \citep{cao2026tsmrg}.

Although the aforementioned works have significantly enriched cross-modal alignment methods, they mostly focus on reducing the relative distance between instance samples or rely on attention-based feature fusion. These strategies fail to effectively eliminate the inherent modality gap, resulting in a persistent distribution shift within the image-text latent space. Therefore, these methods struggle to extract purified features as semantic anchors accurately aligned with diseases, leaving the model without a solid diagnostic foundation when subsequently processing heterogeneous clinical evidence.

\subsection{Heterogeneous Evidence Scheduling}

To enrich diagnostic details in reports, recent studies have extensively introduced multi-source prior knowledge and heterogeneous clinical evidence. Early methods, such as PPKED \citep{liu2021exploring} and GSKET \citep{yang2022knowledge}, provided global guidance using medical knowledge graphs. To enhance semantic support and feature interactions, MSAT \citep{wang2022medical} incorporated an external medical concept network, while XProNet \citep{wang2022cross} learned cross-modal prototypes. To further improve evidence aggregation, recent architectures utilize tree-graph reasoning \citep{hou2023organ}, expert token mechanisms \citep{wang2023metransformer}, and dual-modality multi-scale visual feature flows \citep{tang2025dual}. To directly guide the decoding process, BootstrappingLLM \citep{liu2024bootstrapping} and DDaTR \citep{song2025ddatr} employ retrieved instances and longitudinal information, respectively. To inject richer contextual information into LLMs, recent approaches introduce contextual residuals \citep{wang2026r2gencsr}, construct associative memory modules via Hopfield networks \citep{wang2026activating}, and utilize dual-modality similarity retrieval mechanisms \citep{wu2026disease}. Additionally, FedMRG \citep{che2026llm} employs dual adapters to handle heterogeneous reporting styles within a federated learning framework.

Although these works have significantly expanded the prior contexts of models, they often lack dynamic and targeted semantic reconstruction for each type of heterogeneous information when processing highly heterogeneous clinical evidence composed of visual features, local entities, and global retrievals. Furthermore, during the subsequent scheduling of multi-source evidence, these methods typically adopt predefined aggregation paradigms, failing to adaptively allocate fusion weights based on the weakness of specific pathological features. This undifferentiated rigid coupling mechanism allows massive text priors and the inherent generation inertia of LLMs to dominate the autoregressive decoding process. As a result, weak visual abnormality cues are diluted, which ultimately leads to templated report expressions and makes it difficult to accurately depict complex imaging details.

\begin{figure*}[pos=tb]
    \centering
    \includegraphics[width=\linewidth]{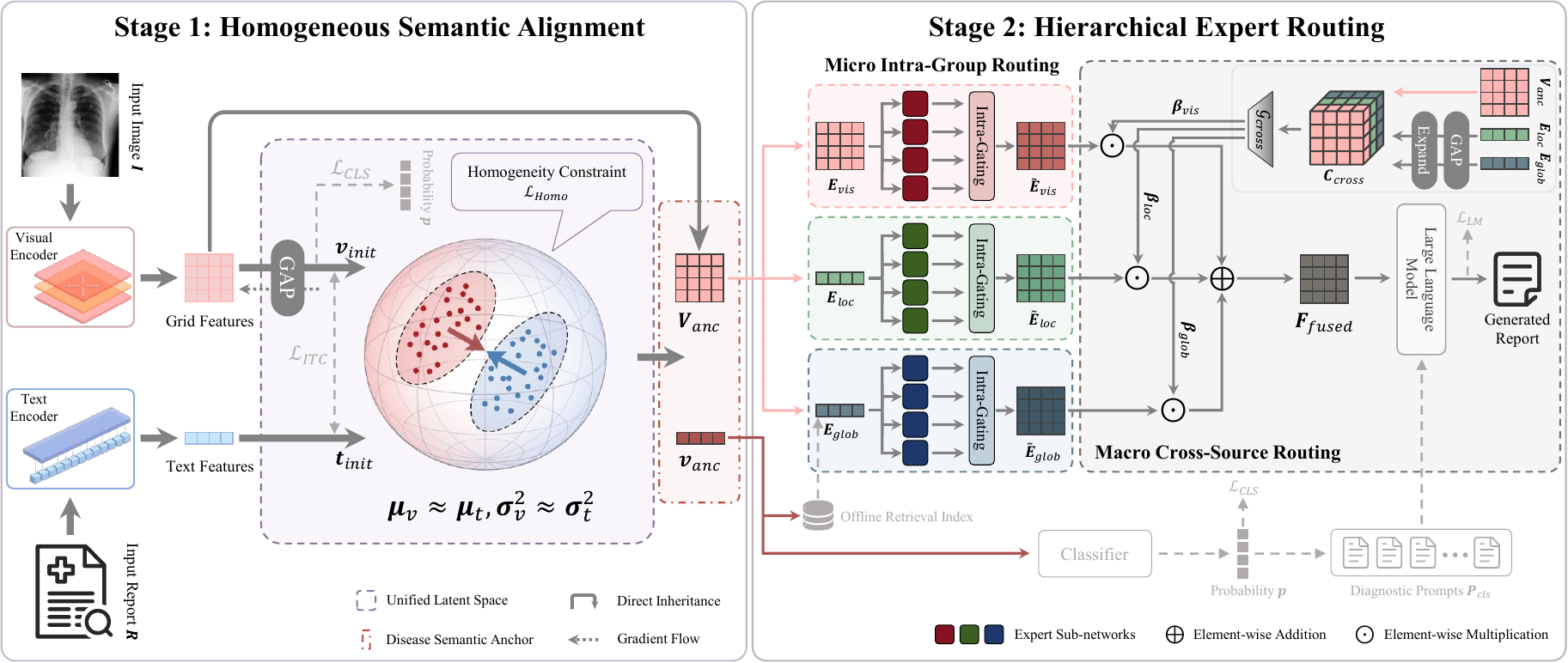}
    \caption{Overall architecture of the proposed HSA-HER framework. \textbf{(Stage 1)} The model eliminates distribution shifts in a unified latent space via an explicit homogeneous distribution constraint to establish precise disease semantic anchors. \textbf{(Stage 2)} Guided by these anchors, micro intra-group routing activates expert sub-networks to perform targeted reconstruction on heterogeneous evidence. Macro cross-source routing adaptively allocates fusion weights and outputs the final diagnostic evidence to drive the large language model for decoding.}
    \label{FIG:2}
\end{figure*}

\section{Method}

In this section, we detail the proposed two-stage homogeneous semantic alignment and hierarchical expert routing (HSA-HER) framework for radiology report generation.

\subsection{Task Definition and Overall Architecture}

Radiology report generation (RRG) is fundamentally a cross-modal conditional sequence generation problem. Given a patient's chest radiograph $\boldsymbol{I} \in \mathbb{R}^{C \times H \times W}$, the model aims to generate a clinical diagnostic report $\boldsymbol{R} = \{y_1, y_2, \dots, y_L\}$ comprising $L$ tokens. Under the autoregressive generation paradigm, given the parameters $\theta$, the probability of generating the report $\boldsymbol{R}$ can be factored into the product of conditional probabilities at each time step:
\begin{equation}
p(\boldsymbol{R}\vert{}\boldsymbol{I}; \theta) = \prod_{t=1}^{L} p(y_t \vert{} y_{<t}, \boldsymbol{I}; \theta)
\label{eq:autoregressive}
\end{equation}
where $y_{<t}$ denotes the historical tokens generated before time step $t$. To maximize this probability and accurately depict weak pathological details, we propose a computational pipeline that completely decouples feature alignment and information scheduling.

As illustrated in Figure \ref{FIG:2}, HSA-HER is physically decoupled into two core stages. First, the model eliminates cross-modal distribution shifts through an explicit homogeneous distribution constraint, establishing global and dense disease semantic anchors (detailed in Section \ref{Section 3.2}). Subsequently, guided by these anchors, the model deploys a hierarchical expert routing mechanism to perform targeted reconstruction and adaptive aggregation on heterogeneous clinical evidence, ultimately driving the large language model to complete decoding (detailed in Sections \ref{Section 3.3} and \ref{Section 3.4}).

\subsection{Stage 1: Homogeneous Semantic Alignment}
\label{Section 3.2}

\subsubsection{Multi-Modal Feature Initialization}

Given a chest radiograph $\boldsymbol{I} \in \mathbb{R}^{C \times H \times W}$ and the corresponding expert reference diagnostic report $\boldsymbol{R}$, we employ a dual-stream encoder to extract initial multi-modal representations.

In the visual branch, we adopt VMamba to capture long-range dependencies and local textures. The visual encoder processes $\boldsymbol{I}$ into spatial grid features. Global average pooling (GAP) and a visual projection layer then map these features into a unified $d$-dimensional space, yielding the initial visual feature $\boldsymbol{v}_{init} \in \mathbb{R}^{d}$.

In the text branch, we employ Bio\_ClinicalBERT to parse medical terminology and clinical logic. The text encoder processes the tokenized report $\boldsymbol{R}$, extracting the global aggregation token feature. An independent text projection layer then maps this representation into the same $d$-dimensional space, yielding the initial text feature $\boldsymbol{t}_{init} \in \mathbb{R}^{d}$.

Despite sharing the same dimensionality, $\boldsymbol{v}_{init}$ and $\boldsymbol{t}_{init}$ reside in different distribution manifolds due to the inherent modality gap between low-level visual pixels and high-level textual semantics. This necessitates the subsequent deep homogeneous alignment.

\subsubsection{Explicit Homogeneous Distribution Constraint}

To break the inherent heterogeneity barrier between visual and textual features, we introduce an explicit homogeneous distribution constraint mechanism after feature initialization. This mechanism aims to force the image and text representations to follow a consistent statistical distribution in a unified latent space, thereby eliminating cross-modal distribution shifts. 

Given a batch of initial visual features $\boldsymbol{v}_{init}$ and text features $\boldsymbol{t}_{init}$ with a batch size of $B$, the model first performs normalization along the feature dimension. Specifically, it projects them onto a unit hypersphere to obtain the normalized features $\hat{\boldsymbol{v}}$ and $\hat{\boldsymbol{t}}$. Subsequently, the model calculates the first-order and second-order statistical moments of the visual and text features along the batch dimension, namely the mean vectors $\boldsymbol{\mu}_v, \boldsymbol{\mu}_t$ and the uncorrected variance vectors $\boldsymbol{\sigma}^2_v, \boldsymbol{\sigma}^2_t$. These complementary statistics jointly characterize the central tendency and dispersion of each modality in the normalized latent space, providing a compact distribution-level description for subsequent alignment. The mathematical calculations are as follows:
\begin{equation}
\boldsymbol{\mu}_v = \frac{1}{B} \sum_{i=1}^{B} \hat{\boldsymbol{v}}_i, \quad \boldsymbol{\mu}_t = \frac{1}{B} \sum_{i=1}^{B} \hat{\boldsymbol{t}}_i
\end{equation}
\begin{equation}
\boldsymbol{\sigma}^2_v = \frac{1}{B} \sum_{i=1}^{B} (\hat{\boldsymbol{v}}_i - \boldsymbol{\mu}_v)^2, \quad \boldsymbol{\sigma}^2_t = \frac{1}{B} \sum_{i=1}^{B} (\hat{\boldsymbol{t}}_i - \boldsymbol{\mu}_t)^2
\end{equation}

To achieve deep semantic homogenization, the model constructs the homogeneous distribution constraint loss $\mathcal{L}_{Homo}$ by minimizing the squared Euclidean distances between the corresponding cross-modal statistical moments:
\begin{equation}
\mathcal{L}_{Homo} = \vert{}\vert{}\boldsymbol{\mu}_v - \boldsymbol{\mu}_t\vert{}\vert{}_2^2 + \vert{}\vert{}\boldsymbol{\sigma}^2_v - \boldsymbol{\sigma}^2_t\vert{}\vert{}_2^2
\end{equation}

Through the aforementioned distribution-level constraint, the model effectively reduces the distance between the overall distribution manifolds of different modalities in the low-level feature space. Consequently, this establishes a mathematical foundation for subsequently establishing precise disease semantic anchors.

\subsubsection{Establishment of Disease Semantic Anchors}

The homogeneous distribution constraint establishes a solid global statistical foundation for cross-modal alignment. Building upon this, to further capture fine-grained pathological semantics and explicitly inject clinical priors, we introduce image-text contrastive learning and a logit-adjusted multi-label disease classifier, ultimately establishing precise disease semantic anchors.

First, to align positive image-text pairs and separate negative samples, we apply the standard InfoNCE loss $\mathcal{L}_{ITC}$ \citep{radford2021learning} to the global visual and textual features. Second, medical imaging datasets inherently suffer from long-tailed label distributions. Relying solely on contrastive learning can bias the representation space of rare lesions toward majority classes. Therefore, we append a multi-label disease classifier to the visual branch and compute a logit-adjusted binary cross-entropy loss $\mathcal{L}_{CLS}$ \citep{menon2021long}. By applying a prior frequency penalty $-\log \pi_c$ to the original prediction logits, this adjustment explicitly suppresses the confidence of high-frequency classes. Consequently, it forces the model to retain sensitivity to subtle minority pathological features in its low-level representations.

The total first-stage optimization objective $\mathcal{L}_{stage1}$ is detailed in Section \ref{Section 3.4}. Jointly optimized by $\mathcal{L}_{Homo}$, $\mathcal{L}_{ITC}$, and $\mathcal{L}_{CLS}$, the visual features eliminate modality heterogeneity and integrate fine-grained pathological semantics with balanced decision boundaries. We define this optimized visual space as the disease semantic anchor. To accommodate multi-dimensional decoding, it is physically decoupled into two forms. The globally pooled global semantic anchor $\boldsymbol{v}_{anc}$ interfaces with external memory retrieval and diagnostic prompt generation. Meanwhile, the dense semantic anchor $\boldsymbol{V}_{anc}$ completely retains 2D spatial topology, serving directly as the query for subsequent hierarchical expert routing. This provides foundational support for the targeted reconstruction and dynamic weighting of heterogeneous evidence.

\subsection{Stage 2: Hierarchical Expert Routing}
\label{Section 3.3}

\subsubsection{Heterogeneous Clinical Evidence Construction}
\label{Section 3.3.1}

The established semantic anchors provide precise disease guidance. Guided by these anchors, the model further integrates fine-grained anatomical morphology and clinical contexts for report generation. Before decoding, the framework injects priors and constructs evidence via two parallel branches.

First, the global semantic anchor $\boldsymbol{v}_{anc}$ is fed into the Stage 1 multi-label classifier. The resulting prediction probability $\boldsymbol{p} \in \mathbb{R}^C$ is discretized into a structured diagnostic prompt sequence $\boldsymbol{P}_{cls}$. Bypassing low-level routing, this prompt is directly injected into the large language model's conditional context as an explicit prior to establish a macro diagnostic tone.

Second, the model constructs a heterogeneous clinical evidence set comprising three components. The visual feature $\boldsymbol{E}_{vis} \in \mathbb{R}^{N \times d}$ is inherited directly from the dense semantic anchor ($\boldsymbol{E}_{vis} = \boldsymbol{V}_{anc}$). It retains the 2D spatial topology and local pathological textures of the original radiograph, serving as the foundational morphological evidence for self-reconstruction. The local entity cue $\boldsymbol{E}_{loc} \in \mathbb{R}^{L_{loc} \times d}$ is constructed from entity-level representations extracted by MedKLIP \citep{wu2023medklip}. For each radiograph, MedKLIP provides representations of the spatial locations and presence states of 75 predefined medical entities, which are precomputed and stored as \texttt{.npy} files. These static embeddings provide lesion-level and anatomy-level cues. The global retrieval prior $\boldsymbol{E}_{glob} \in \mathbb{R}^{L_{glob} \times d}$ provides a reference for global diagnostic semantics and report expression patterns with zero online search overhead. It is formed by directly loading the static text representations of the top-$K$ ($K=5$) reference reports from the training set, which are retrieved offline using $\boldsymbol{v}_{anc}$ as the query vector. Decoupled in information granularity and modality, these three features ($\boldsymbol{E}_{vis}$, $\boldsymbol{E}_{loc}$, $\boldsymbol{E}_{glob}$) constitute the heterogeneous evidence space for subsequent hierarchical expert routing.

\subsubsection{Micro Intra-Group Targeted Reconstruction}

After obtaining the heterogeneous clinical evidence set composed of $\boldsymbol{E}_{vis}$, $\boldsymbol{E}_{loc}$, and $\boldsymbol{E}_{glob}$, directly applying simple feature concatenation or channel addition triggers spatial misalignment and the dilution of visual cues. To address this, we deploy a micro intra-group routing mechanism within each heterogeneous evidence branch. Using the dense semantic anchor $\boldsymbol{V}_{anc}$ established in Stage 1 as the query condition, this mechanism performs targeted mining and semantic reconstruction on the feature space of each branch.

For any evidence branch $x \in \{vis, loc, glob\}$, let its internal preset expert sub-network cluster be $\Phi_{x} = \{\phi_{x,1}, \dots, \phi_{x,M}\}$, where $M$ is the number of experts in this branch (set to $M=4$ in the experiments). Physically, each expert sub-network $\phi_{x,m}(\cdot)$ is a Transformer cross-attention module. The model forcibly assigns the dense semantic anchor $\boldsymbol{V}_{anc}$ as the query vector, and assigns the clinical evidence $\boldsymbol{E}_x$ of the corresponding branch as the key and value. Note that for the visual branch ($x=vis$), since the sources of the query and the key-value are physically identical ($\boldsymbol{V}_{anc} = \boldsymbol{E}_{vis}$), this operation naturally degenerates into a self-attention mechanism. Specifically, the targeted mining process of the $m$-th expert network can be strictly defined as:
\begin{equation}
\hat{\boldsymbol{H}}_{x,m} = \boldsymbol{V}_{anc} + \text{MHA}_{x,m}\bigl(\text{LN}(\boldsymbol{V}_{anc}), \text{LN}(\boldsymbol{E}_x), \text{LN}(\boldsymbol{E}_x)\bigr)
\end{equation}
\begin{equation}
\boldsymbol{H}_{x,m} = \hat{\boldsymbol{H}}_{x,m} + \text{MLP}_{x,m}\bigl(\text{LN}(\hat{\boldsymbol{H}}_{x,m})\bigr)
\end{equation}
where $\text{LN}(\cdot)$ denotes layer normalization, while $\text{MHA}_{x,m}(\cdot)$ and $\text{MLP}_{x,m}(\cdot)$ denote the expert-specific multi-head attention and feed-forward modules of the $m$-th expert in branch $x$, respectively. The resulting $\boldsymbol{H}_{x,m} \in \mathbb{R}^{N \times d}$ is the specialized clinical representation output by this expert after reconstruction. This design ensures that the information extracted from the heterogeneous clinical evidence can be strictly reconstructed under the guidance of the spatial topology encoded in \(\boldsymbol{V}_{anc}\).

\begin{figure}[pos=t]
    \centering
    \includegraphics[width=\linewidth]{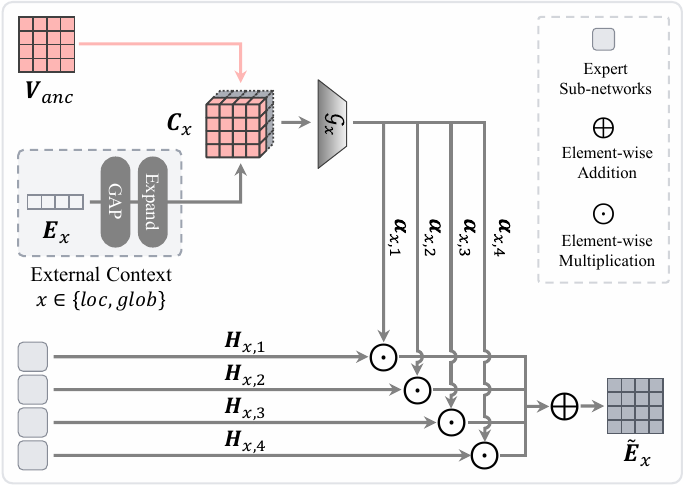}
    \caption{Detailed architecture of intra-group gating. Guided by the context $\boldsymbol{C}_x$, the gating network dynamically allocates weights to aggregate expert outputs $\boldsymbol{H}_{x,m}$, yielding the reconstructed representation $\tilde{\boldsymbol{E}}_x$.}
    \label{FIG:3}
\end{figure}

After obtaining the independent outputs of all experts, we introduce an intra-group gating network $\mathcal{G}_x(\cdot)$ to determine the contribution ratio of each expert, as illustrated in Figure \ref{FIG:3}. To endow the gating decision with fine-grained spatial awareness, the feature context $\boldsymbol{C}_x$ input to the gate adopts a differentiated construction strategy. For the visual branch ($x=vis$), the dense semantic anchor $\boldsymbol{V}_{anc}$ directly serves as the gating input. However, for the local and global external evidence branches ($x \in \{loc, glob\}$), it is formed by concatenating $\boldsymbol{V}_{anc}$ and the spatially expanded GAP context feature of the current evidence branch along the channel dimension. Subsequently, the gating network calculates the dynamic activation weight $\boldsymbol{\alpha}_x \in \mathbb{R}^{N \times M}$ for each expert:
\begin{equation}
\boldsymbol{\alpha}_x = \text{Softmax}(\mathcal{G}_x(\boldsymbol{C}_x))
\end{equation}

Finally, the features extracted by each expert are dynamically weighted and aggregated within the group to generate the targeted output $\tilde{\boldsymbol{E}}_x$ reconstructed at the micro level for this branch:
\begin{equation}
\tilde{\boldsymbol{E}}_x = \sum_{m=1}^M \boldsymbol{\alpha}_{x,m} \odot \boldsymbol{H}_{x,m}
\end{equation}

Through the aforementioned micro intra-group reconstruction operation, the originally loose and noisy heterogeneous features are uniformly projected into the diagnostic space centered on the dense visual semantic distribution. Consequently, this forms three highly specialized purified representations $\tilde{\boldsymbol{E}}_{vis}$, $\tilde{\boldsymbol{E}}_{loc}$, and $\tilde{\boldsymbol{E}}_{glob}$, thereby effectively eliminating interference for the top-level cross-source fusion.

\subsubsection{Macro Cross-Source Adaptive Fusion}

Through micro intra-group reconstruction, the model obtains three highly specialized feature representations: $\tilde{\boldsymbol{E}}_{vis}$, $\tilde{\boldsymbol{E}}_{loc}$, and $\tilde{\boldsymbol{E}}_{glob}$. Considering the dynamic variations in the reliance on different types of evidence during complex imaging diagnosis, we introduce a macro cross-source routing mechanism at the top level to achieve the adaptive and structured fusion of multi-source clinical evidence.

First, the model constructs a cross-source feature context $\boldsymbol{C}_{cross}$ to guide the top-level routing decision. To ensure that the decision accounts for both the fine-grained spatial topology at the low level and the global context of external evidence, $\boldsymbol{C}_{cross}$ is formed by concatenating the dense semantic anchor $\boldsymbol{V}_{anc}$ and the spatially expanded GAP contexts of the two external evidence branches along the channel dimension:
\begin{equation}
\boldsymbol{C}_{cross} = [\boldsymbol{V}_{anc}, \bar{\boldsymbol{E}}_{loc}, \bar{\boldsymbol{E}}_{glob}] \in \mathbb{R}^{N \times 3d}
\end{equation}
where $\bar{\boldsymbol{E}}_{loc}$ and $\bar{\boldsymbol{E}}_{glob}$ represent the context features after masked GAP and expansion along the spatial dimension $N$.

Subsequently, the model deploys a top-level gating network $\mathcal{G}_{cross}(\cdot)$ to calculate the cross-source fusion weights. Considering that external heterogeneous evidence may be missing or invalid in extreme cases, we introduce an allocation function with a validity mask mechanism to ensure the numerical stability of the network. Let $\boldsymbol{M}_{valid} \in \{0, 1\}^3$ be a boolean mask indicating whether the evidence of each branch is valid. The allocation weights $\boldsymbol{\beta}^{(n)} \in \mathbb{R}^3$ for the three branches at the $n$-th spatial position are calculated as follows:
\begin{equation}
\boldsymbol{\beta}^{(n)} = \text{MaskedSoftmax}(\mathcal{G}_{cross}(\boldsymbol{C}_{cross}^{(n)}), \boldsymbol{M}_{valid})
\end{equation}
This mask mechanism forcibly blocks the gradient and weight allocation for invalid evidence branches. Therefore, it ensures that the model safely falls back to relying on its own visual evidence when external priors are missing.

Finally, at the token level of the dense semantic anchor, the model adaptively weights and fuses the three features to generate the final structured multi-modal diagnostic evidence $\boldsymbol{F}_{fused} \in \mathbb{R}^{N \times d}$:
\begin{equation}
\boldsymbol{F}_{fused} = \boldsymbol{\beta}_{vis} \odot \tilde{\boldsymbol{E}}_{vis} + \boldsymbol{\beta}_{loc} \odot \tilde{\boldsymbol{E}}_{loc} + \boldsymbol{\beta}_{glob} \odot \tilde{\boldsymbol{E}}_{glob}
\end{equation}

Through the complete computational pipeline of homogeneous semantic alignment, micro intra-group reconstruction, and macro cross-source fusion, $\boldsymbol{F}_{fused}$ effectively eliminates the interference of modality heterogeneity and deeply integrates visual morphology with cross-source clinical priors. Thereafter, this feature, together with the explicit diagnostic prompts $\boldsymbol{P}_{cls}$ constructed in Section \ref{Section 3.3.1}, is fed into the large language model to drive the subsequent high-precision autoregressive text decoding.

\subsection{Training Objectives}
\label{Section 3.4}

The training process is decoupled into two stages. 

In the first stage, to achieve homogeneous alignment of low-level representations, we pre-train the dual-stream encoder by jointly optimizing the homogeneous distribution constraint loss $\mathcal{L}_{Homo}$, the contrastive loss $\mathcal{L}_{ITC}$, and the classification loss $\mathcal{L}_{CLS}$:
\begin{equation}
\mathcal{L}_{stage1} = \mathcal{L}_{Homo} + \mathcal{L}_{ITC} + \mathcal{L}_{CLS}
\end{equation}

In the second stage, the encoder is initialized with the pre-trained weights. The fused feature $\boldsymbol{F}_{fused}$ and the discretized explicit diagnostic prompts $\boldsymbol{P}_{cls}$ are fed into the large language model for autoregressive decoding. Given a ground-truth expert report $\boldsymbol{R} = \{y_1, y_2, \dots, y_L\}$ consisting of $L$ tokens, text generation is optimized using the negative log-likelihood loss $\mathcal{L}_{LM}$:
\begin{equation}
\mathcal{L}_{LM} = - \sum_{t=1}^{L} \log p(y_t \vert{} y_{<t}, \boldsymbol{F}_{fused}, \boldsymbol{P}_{cls}; \theta_{LLM})
\end{equation}
where $\theta_{LLM}$ represents the LLM parameters, and $y_{<t}$ denotes the tokens generated before time step $t$.

To continuously impose low-level pathological constraints during the decoding phase, we combine the disease classification loss $\mathcal{L}_{CLS}$ from the first stage with $\mathcal{L}_{LM}$ to form the second-stage objective $\mathcal{L}_{stage2}$:
\begin{equation}
\mathcal{L}_{stage2} = \mathcal{L}_{LM} + \lambda_{cls} \mathcal{L}_{CLS}
\end{equation}
where $\lambda_{cls}$ is a hyperparameter balancing textual fluency and pathological discrimination. 

Driven by this joint optimization strategy, HSA-HER effectively prevents the dilution of weak visual cues caused by autoregressive generation inertia and dominant text priors, ensuring the accurate depiction of complex pathological details in the generated reports.

\begin{table*}[t]
    \caption{Performance comparison on the CheXpert Plus test set. The best results are highlighted in bold.}
    \label{tab:main_results_chexplus}
    \centering 
    \small     
    \resizebox{\textwidth}{!}{
    \begin{tabular}{llccccccccc}
        \toprule
        \multirow{2}{*}{\raisebox{-0.6ex}{Method}} & \multirow{2}{*}{\raisebox{-0.6ex}{Publication}} & \multicolumn{6}{c}{NLG Metrics} & \multicolumn{3}{c}{CE Metrics} \\
        \cmidrule(lr){3-8} \cmidrule(lr){9-11}
        & & BLEU-1 & BLEU-2 & BLEU-3 & BLEU-4 & ROUGE-L & CIDEr & Precision & Recall & F1 \\
        \midrule
        R2Gen \cite{chen2020generating} & EMNLP'20 & 0.301 & 0.179 & 0.118 & 0.081 & 0.246 & 0.077 & 0.318 & 0.200 & 0.181 \\
        WCL \cite{yan2021weakly} & EMNLP'21 & -- & -- & -- & 0.084 & 0.253 & 0.103 & 0.335 & 0.259 & 0.256 \\
        R2GenCMN \cite{chen2021cross} & ACL'21 & 0.321 & 0.195 & 0.128 & 0.087 & 0.256 & 0.102 & 0.329 & 0.241 & 0.231 \\
        XProNet \cite{wang2022cross} & ECCV'22 & 0.364 & 0.225 & 0.148 & 0.100 & 0.265 & 0.121 & 0.314 & 0.247 & 0.259 \\
        MSAT \cite{wang2022medical} & MICCAI'22 & -- & -- & -- & 0.036 & 0.156 & 0.018 & 0.044 & 0.142 & 0.057 \\
        R2GenRL \cite{qin2022reinforced} & ACL'22 & -- & -- & -- & 0.035 & 0.186 & 0.012 & 0.193 & 0.229 & 0.196 \\
        TIMER \cite{wu2023token} & CHIL'23 & -- & -- & -- & 0.083 & 0.254 & 0.104 & 0.345 & 0.238 & 0.234 \\
        CAMANet \cite{wang2024camanet} & JBHI'23 & -- & -- & -- & 0.083 & 0.249 & 0.090 & 0.328 & 0.224 & 0.216 \\
        M2KT \cite{yang2023radiology} & MedIA'23 & -- & -- & -- & 0.078 & 0.247 & 0.077 & 0.044 & 0.142 & 0.058 \\
        R2GenGPT \cite{wang2023r2gengpt} & Meta-Rad'23 & 0.361 & 0.224 & 0.149 & 0.101 & 0.266 & 0.123 & 0.315 & 0.244 & 0.260 \\
        ORGan \cite{hou2023organ} & ACL'23 & 0.320 & 0.196 & 0.128 & 0.086 & 0.261 & 0.107 & 0.288 & 0.287 & 0.277 \\
        ASGMD \cite{xue2024generating} & ESWA'24 & 0.267 & 0.149 & 0.094 & 0.063 & 0.220 & 0.044 & -- & -- & -- \\
        Token-Mixer \cite{yang2024token} & TMI'24 & 0.378 & 0.231 & 0.153 & 0.091 & 0.262 & 0.098 & 0.309 & 0.270 & 0.288 \\
        PromptMRG \cite{jin2024promptmrg} & AAAI'24 & 0.326 & 0.174 & -- & 0.095 & 0.222 & 0.044 & 0.258 & 0.265 & 0.281 \\
        VLCI \cite{chen2025cross} & TIP'25 & -- & -- & -- & 0.080 & 0.247 & 0.072 & 0.341 & 0.175 & 0.163 \\
        MambaXray \cite{wang2025cxpmrg} & CVPR'25 & -- & -- & -- & \textbf{0.112} & 0.276 & 0.139 & 0.377 & 0.319 & 0.335 \\
        MCA-RG \cite{xing2025mca} & MICCAI'25 & 0.367 & 0.218 & 0.149 & 0.102 & 0.266 & -- & -- & -- & -- \\
        R2GenCSR \cite{wang2026r2gencsr} & JBHI'26 & 0.364 & 0.225 & 0.148 & 0.100 & 0.265 & 0.121 & -- & -- & -- \\
        AM-MRG \cite{wang2026activating} & TMI'26 & 0.381 & 0.238 & 0.159 & 0.109 & 0.282 & \textbf{0.221} & \textbf{0.396} & 0.318 & 0.336 \\
        \textbf{HSA-HER} & \textbf{Ours} & \textbf{0.383} & \textbf{0.240} & \textbf{0.161} & 0.111 & \textbf{0.283} & 0.212 & 0.378 & \textbf{0.325} & \textbf{0.338} \\
        \bottomrule
    \end{tabular}
    }
    \vskip -0.1in
\end{table*}

\section{Experiments}

\subsection{Datasets}

We evaluate HSA-HER on three widely used public chest X-ray datasets, namely CheXpert Plus \citep{chambon2024chexpert}, IU X-Ray \citep{demnerfushman2016preparing}, and MIMIC-CXR \citep{johnson2019mimic}. 

The \textbf{CheXpert Plus} dataset comprises 223,228 chest radiographs paired with 187,711 anonymized clinical reports from 64,725 patients, featuring 14 chest pathology labels and RadGraph annotations \citep{jain2021radgraph}. The \textbf{IU X-Ray} dataset, provided by Indiana University, includes 7,470 frontal and lateral chest X-ray images associated with 3,955 diagnostic reports. The \textbf{MIMIC-CXR} dataset is the most extensive public dataset containing paired chest X-ray images and radiology reports, consisting of 377,110 images and 227,835 diagnostic reports from 65,379 patients. For MIMIC-CXR, we strictly follow the standard split and preprocessing pipeline established by \citet{chen2020generating}, resulting in 270,790 training, 2,130 validation, and 3,858 testing samples.

\begin{table*}[t]
    \caption{Performance comparison on the IU X-Ray and MIMIC-CXR test sets. The best results are highlighted in bold.}
    \label{tab:main_results_iu_mimic}
    \centering
    \small
    
    \resizebox{\textwidth}{!}{
    \begin{tabular}{llcccccc} 
        \toprule
        Method & Publication & BLEU-1 & BLEU-2 & BLEU-3 & BLEU-4 & ROUGE-L & CIDEr \\
        \midrule
        
        \rowcolor{gray!20} \multicolumn{8}{l}{\textbf{IU X-Ray}} \\
        
        R2Gen \cite{chen2020generating} & EMNLP'20 & 0.470 & 0.304 & 0.219 & 0.165 & 0.371 & -- \\
        PPKED \cite{liu2021exploring} & CVPR'21 & 0.483 & 0.315 & 0.224 & 0.168 & 0.376 & 0.351 \\
        CMCL \cite{liu2021competence} & ACL'21 & 0.473 & 0.305 & 0.217 & 0.162 & 0.378 & -- \\
        Clinical-BERT \cite{yan2022clinical} & AAAI'22 & 0.495 & 0.330 & 0.231 & 0.170 & 0.376 & 0.432 \\
        GSKET \cite{yang2022knowledge} & MedIA'22 & 0.496 & 0.327 & 0.238 & 0.178 & 0.381 & 0.382 \\
        METransformer \cite{wang2023metransformer} & CVPR'23 & 0.483 & 0.322 & 0.228 & 0.172 & 0.380 & 0.435 \\
        DCL \cite{li2023dynamic} & CVPR'23 & -- & -- & -- & 0.163 & 0.383 & 0.586 \\
        R2GenGPT \cite{wang2023r2gengpt} & Meta-Rad'23 & 0.465 & 0.299 & 0.214 & 0.161 & 0.376 & 0.542 \\
        PromptMRG \cite{jin2024promptmrg} & AAAI'24 & 0.401 & -- & -- & 0.098 & 0.160 & -- \\
        DDaTR \cite{song2025ddatr} & TMI'25 & 0.421 & -- & -- & 0.103 & 0.307 & -- \\
        MambaXray \cite{wang2025cxpmrg} & CVPR'25 & 0.491 & 0.330 & 0.241 & 0.185 & 0.371 & 0.524 \\
        DADS \cite{wu2026disease} & AAAI'26 & 0.495 & 0.328 & 0.242 & \textbf{0.187} & 0.384 & \textbf{0.634} \\
        FedMRG \cite{che2026llm} & TMI'26 & 0.436 & 0.264 & 0.173 & 0.120 & 0.297 & -- \\
        \textbf{HSA-HER} & \textbf{Ours} & \textbf{0.502} & \textbf{0.337} & \textbf{0.245} & 0.186 & \textbf{0.385} & 0.490 \\
        
        \midrule
        \rowcolor{gray!20} \multicolumn{8}{l}{\textbf{MIMIC-CXR}} \\
        
        R2Gen \cite{chen2020generating} & EMNLP'20 & 0.353 & 0.218 & 0.145 & 0.103 & 0.277 & -- \\
        PPKED \cite{liu2021exploring} & CVPR'21 & 0.360 & 0.224 & 0.149 & 0.106 & 0.284 & 0.237 \\
        CMCL \cite{liu2021competence} & ACL'21 & 0.344 & 0.217 & 0.140 & 0.097 & 0.281 & -- \\
        Clinical-BERT \cite{yan2022clinical} & AAAI'22 & 0.383 & 0.230 & 0.151 & 0.106 & 0.275 & 0.151 \\
        GSKET \cite{yang2022knowledge} & MedIA'22 & 0.363 & 0.228 & 0.156 & 0.115 & 0.284 & 0.203 \\
        M2KT \cite{yang2023radiology} & MedIA'23 & 0.386 & 0.237 & 0.157 & 0.111 & 0.274 & 0.111 \\
        BootstrappingLLM \cite{liu2024bootstrapping} & AAAI'24 & 0.402 & 0.262 & 0.180 & 0.128 & \textbf{0.291} & -- \\
        PromptMRG \cite{jin2024promptmrg} & AAAI'24 & 0.398 & -- & -- & 0.112 & 0.268 & -- \\
        DACG \cite{lang2025dacg} & MedIA'24 & 0.398 & 0.249 & 0.167 & 0.117 & 0.290 & -- \\
        DMVF \cite{tang2025dual} & MedIA'24 & 0.396 & 0.245 & 0.162 & 0.116 & 0.289 & 0.116 \\
        VLCI \cite{chen2025cross} & TIP'25 & 0.400 & 0.245 & 0.165 & 0.119 & 0.280 & -- \\
        DDaTR \cite{song2025ddatr} & TMI'25 & 0.401 & -- & -- & 0.113 & 0.272 & -- \\
        MambaXray \cite{wang2025cxpmrg} & CVPR'25 & 0.422 & 0.268 & 0.184 & 0.133 & 0.289 & 0.241 \\
        FedMRG \cite{che2026llm} & TMI'26 & 0.402 & 0.255 & 0.176 & 0.127 & 0.276 & -- \\
        TSMRG \cite{cao2026tsmrg} & TMM'26 & 0.403 & 0.244 & 0.161 & 0.113 & 0.272 & -- \\
        \textbf{HSA-HER} & \textbf{Ours} & \textbf{0.422} & \textbf{0.269} & \textbf{0.185} & \textbf{0.135} & 0.288 & \textbf{0.248} \\
        \bottomrule
    \end{tabular}
    }
    \vskip -0.1in
\end{table*}

\subsection{Evaluation metrics}
To comprehensively evaluate the performance of our model, we employ natural language generation (NLG) metrics, clinical efficacy (CE) metrics, and the GREEN diagnostic safety metric.

For the NLG metrics, we report BLEU-1 to BLEU-4 \citep{papineni2002bleu}, ROUGE-L \citep{lin2004rouge}, and CIDEr \citep{vedantam2015cider}. These metrics characterize the consistency between the generated texts and the ground-truth reports from complementary perspectives: BLEU measures modified $n$-gram precision, ROUGE-L captures sequence-level similarity based on the longest common subsequence, and CIDEr measures consensus using TF-IDF-weighted $n$-grams.

For the CE metrics, we follow the evaluation protocol adopted by \citet{wang2026activating}. We use CheXbert \citep{smit2020chexbert} to extract clinical observation labels from both the predicted and ground-truth reports and compare their presence to assess diagnostic accuracy. Specifically, we report macro-averaged Precision, Recall, and F1 scores across the evaluated clinical observation categories.

Finally, considering the unique characteristics of radiology reports, we further adopt the GREEN \citep{ostmeier2024green} metric to evaluate diagnostic safety. This framework utilizes a language model to conduct a fine-grained evaluation of clinical consistency between the generated and reference reports, explicitly distinguishing between clinically significant and insignificant errors. The GREEN score strictly penalizes six categories of clinically significant errors that may influence clinical decisions (specific error categories are defined in Table \ref{tab:green_metrics}). For the $i$-th sample, its GREEN score is defined as the ratio of the number of matched clinical findings to the sum of this number and the total count of the six clinically significant errors. If no matched findings are present ($\text{\#Matched Findings}_i = 0$), the score is explicitly set to $0$. The calculation is formulated as:
\begin{equation}
\text{GREEN}_i = \frac{\text{\#Matched Findings}_i}{\text{\#Matched Findings}_i + \sum_{j=(a)}^{(f)} \text{\#Error}_{\mathrm{sig},i,j}}
\label{eq:green}
\end{equation}

\subsection{Implementation details}
The visual encoder is instantiated as VMamba \citep{liu2024vmamba}, and the input images are resized to a resolution of $224 \times 224$. The model training was divided into two stages. In the first stage, Bio\_ClinicalBERT \citep{alsentzer2019publicly} was employed as the text encoder. This stage was trained on four NVIDIA A40 GPUs with a batch size of 128 per GPU. In the second stage, the decoder backbone utilized Llama-2-7B-Chat \citep{touvron2023llama} for the CheXpert Plus and MIMIC-CXR datasets, and Qwen1.5-1.8B-Chat \citep{qwen15} for the IU X-Ray dataset. Constrained by the memory overhead of the large language models, the second stage was conducted on a single A40 GPU, and the batch size was adjusted to 8. During this stage, the visual backbone was initialized with the weights from the first stage and updated via full-parameter fine-tuning, while the newly introduced hierarchical expert routing was trained jointly. Meanwhile, the parameters of the language model were strictly frozen to preserve its pre-trained language generation priors. 

Both training stages utilized the AdamW optimizer with an initial learning rate of $1 \times 10^{-4}$, which was decayed to $1 \times 10^{-6}$ following a cosine annealing schedule. To improve memory efficiency, the DeepSpeed \citep{rasley2020deepspeed} framework and BF16 mixed precision were enabled throughout the training process. During the inference phase, text generation was performed using beam search. For both the primary and ablation experiments, the checkpoint achieving the highest combined score of $0.5 \times \text{BLEU-4} + 0.5 \times \text{CIDEr}$ on the validation set was selected for final testing.

\begin{table*}[t]
    \caption{GREEN performance comparison on the MIMIC-CXR and CheXpert Plus test sets. The six clinically significant error categories are defined as follows: (a) False report of a finding, (b) Missing a finding, (c) Misidentification of anatomic location, (d) Misassessment of severity, (e) Mentioning a comparison not in the reference, and (f) Omitting a comparison detailing a change from a prior study. For clinically significant errors (a)--(f) and their total count, lower values are better ($\downarrow$). For matched findings and GREEN score, higher values are better ($\uparrow$).}
    \label{tab:green_metrics}
    \centering
    \setlength{\tabcolsep}{2.3pt}
    \resizebox{\textwidth}{!}{
    \begin{tabular}{llccccccccc}
        \toprule
        \multirow{2}{*}{\raisebox{-0.7ex}{Dataset}} & \multirow{2}{*}{\raisebox{-0.7ex}{Method}} & \multicolumn{6}{c}{\# Clinically Significant Errors $\downarrow$} & \multirow{2}{*}{\raisebox{-0.7ex}{$\sum_{j=a}^{f} \#\mathrm{Error}_{j} \downarrow$}} & \multirow{2}{*}{\raisebox{-0.7ex}{\# Matched Findings $\uparrow$}} & \multirow{2}{*}{\raisebox{-0.7ex}{GREEN $\uparrow$}} \\
        \cmidrule(lr){3-8}
        & & (a) & (b) & (c) & (d) & (e) & (f) & & & \\
        \midrule
        \multirow{4}{*}{MIMIC-CXR} 
        & R2Gen \cite{chen2020generating} & 1.310 & 3.089 & \textbf{0.103} & \textbf{0.201} & 0.082 & 0.142 & 4.926 & 1.803 & 0.283 \\
        & R2GenCMN \cite{chen2021cross} & 1.383 & 2.963 & 0.127 & 0.228 & \textbf{0.081} & 0.161 & 4.942 & 1.935 & 0.297 \\
        & CvT2DistilGPT2 \cite{nicolson2023improving} & \textbf{1.150} & 2.881 & 0.146 & 0.234 & 0.103 & 0.153 & 4.666 & 1.967 & 0.313 \\
        & \textbf{HSA-HER (Ours)} & 1.539 & \textbf{2.317} & 0.161 & 0.288 & 0.150 & \textbf{0.102} & \textbf{4.557} & \textbf{2.066} & \textbf{0.323} \\
        \midrule
        \multirow{4}{*}{CheXpert Plus}
        & R2Gen \cite{chen2020generating} & \textbf{1.235} & 3.638 & \textbf{0.150} & \textbf{0.122} & \textbf{0.021} & 0.103 & 5.269 & 0.938 & 0.168 \\
        & R2GenCMN \cite{chen2021cross} & 1.512 & 3.336 & 0.218 & 0.165 & 0.043 & 0.126 & 5.399 & 1.179 & 0.197 \\
        & CvT2DistilGPT2 \cite{nicolson2023improving} & 1.815 & 3.289 & 0.252 & 0.182 & 0.091 & 0.132 & 5.760 & 1.166 & 0.185 \\
        & \textbf{HSA-HER (Ours)} & 1.960 & \textbf{2.099} & 0.295 & 0.269 & 0.137 & \textbf{0.073} & \textbf{4.832} & \textbf{1.789} & \textbf{0.283} \\
        \bottomrule
    \end{tabular}
    }
\end{table*}

\subsection{Comparison with state-of-the-art methods}
We comprehensively compare the proposed HSA-HER method with state-of-the-art (SOTA) methods for RRG. The comparative evaluations are conducted on the CheXpert Plus, IU X-Ray, and MIMIC-CXR datasets. The reported results of the compared methods are collected either from their original publications or from two recent studies reporting comprehensive comparative results \cite{wu2026disease, wang2026activating}, and all values are used as reported without any modification.

\textbf{Performance on CheXpert Plus.}
As shown in Table \ref{tab:main_results_chexplus}, HSA-HER demonstrates superior performance across both NLG and CE metrics. Regarding the NLG metrics, our model achieves SOTA results in BLEU-1 to BLEU-3 and ROUGE-L, reaching 0.383, 0.240, 0.161, and 0.283, respectively. More importantly, concerning the CE metrics that measure diagnostic accuracy, HSA-HER achieves the highest Recall (0.325) and F1 score (0.338). Furthermore, in the GREEN diagnostic safety evaluation shown in Table \ref{tab:green_metrics}, HSA-HER obtains the highest average matched findings (1.789) and the optimal overall GREEN score (0.283). Specifically, HSA-HER effectively controls Type b and Type f errors, with their average counts reduced to the lowest levels of 2.099 and 0.073, respectively. These results indicate that while generating high-quality reports, HSA-HER effectively improves the coverage of key clinical findings and overall clinical accuracy.

\textbf{Performance on IU X-Ray and MIMIC-CXR.}
Table \ref{tab:main_results_iu_mimic} reports the performance of the model on the IU X-Ray and MIMIC-CXR datasets. On IU X-Ray, HSA-HER achieves the best results in BLEU-1 to BLEU-3 and ROUGE-L. On the largest-scale MIMIC-CXR dataset, HSA-HER achieves SOTA results across all NLG metrics except ROUGE-L. Particularly for the CIDEr metric, which measures consensus using TF-IDF-weighted \(n\)-grams, HSA-HER obtains the highest score of 0.248, while BLEU-4 reaches 0.135. Furthermore, in the GREEN diagnostic safety evaluation shown in Table \ref{tab:green_metrics}, HSA-HER obtains the highest average matched findings (2.066) and the optimal overall GREEN score (0.323) on MIMIC-CXR, with the average counts of Type b and Type f errors reduced to the lowest levels of 2.317 and 0.102, respectively. Notably, Table \ref{tab:green_metrics} shows that HSA-HER has limitations in Type a, c, d, and e errors on CheXpert Plus and MIMIC-CXR. This may reflect the trade-off between improving abnormal finding coverage and calibrating false positives, fine-grained localization, severity, and comparative descriptions, as well as noise and uncertainty propagation from multi-source priors. Overall, this advantage in language generation and clinical safety metrics further demonstrates the capability of HSA-HER to accurately characterize complex imaging details and key diagnostic information.

\begin{table*}[t]
    \caption{Ablation study of Stage 1 and Stage 2 on the CheXpert Plus dataset. Variant \#1 serves as the vanilla baseline. Variants \#2, \#3, \#4, and \#7 progressively validate the Stage 1 constraints ($\mathcal{L}_{ITC}$, $\mathcal{L}_{CLS}$, $\mathcal{L}_{Homo}$). Conversely, Variants \#5, \#6, and \#7 progressively evaluate the Stage 2 modules, namely explicit diagnostic prompts ($\boldsymbol{P}_{cls}$) and hierarchical expert routing (HER), built upon the full Stage 1 setting.}
    \label{tab:ablation_stage1_stage2}
    \centering
    \setlength{\tabcolsep}{2.3pt}
    \resizebox{\textwidth}{!}{
    \begin{tabular}{lcccccccccccccc}
        \toprule
        \multirow{2}{*}{\raisebox{-0.7ex}{Variant}} & \multirow{2}{*}{\raisebox{-0.7ex}{$\mathcal{L}_{ITC}$}} & \multirow{2}{*}{\raisebox{-0.7ex}{$\mathcal{L}_{CLS}$}} & \multirow{2}{*}{\raisebox{-0.7ex}{$\mathcal{L}_{Homo}$}} & \multirow{2}{*}{\raisebox{-0.7ex}{$\boldsymbol{P}_{cls}$}} & \multirow{2}{*}{\raisebox{-0.7ex}{HER}} & \multicolumn{6}{c}{NLG Metrics} & \multicolumn{3}{c}{CE Metrics} \\
        \cmidrule(lr){7-12} \cmidrule(lr){13-15}
        & & & & & & BLEU-1 & BLEU-2 & BLEU-3 & BLEU-4 & ROUGE-L & CIDEr & Precision & Recall & F1 \\
        \midrule
        \#1 &  &  &  &  &  & 0.378 & 0.234 & 0.155 & 0.106 & 0.276 & 0.214 & 0.354 & 0.296 & 0.312 \\
        \#2 &  &  &  & \checkmark & \checkmark & 0.379 & 0.232 & 0.153 & 0.103 & 0.273 & 0.206 & 0.377 & 0.301 & 0.323 \\
        \#3 & \checkmark &  &  & \checkmark & \checkmark & 0.380 & 0.238 & 0.159 & 0.110 & 0.279 & 0.213 & 0.375 & 0.301 & 0.316 \\
        \#4 & \checkmark & \checkmark &  & \checkmark & \checkmark & 0.379 & 0.236 & 0.157 & 0.108 & 0.281 & 0.207 & 0.373 & 0.332 & \textbf{0.345} \\
        \#5 & \checkmark & \checkmark & \checkmark &  &  & 0.383 & 0.237 & 0.157 & 0.107 & 0.279 & \textbf{0.224} & 0.357 & \textbf{0.337} & 0.341 \\
        \#6 & \checkmark & \checkmark & \checkmark & \checkmark &  & 0.381 & 0.237 & 0.158 & 0.108 & 0.281 & 0.218 & 0.378 & 0.329 & 0.344 \\
        \#7 & \checkmark & \checkmark & \checkmark & \checkmark & \checkmark & \textbf{0.383} & \textbf{0.240} & \textbf{0.161} & \textbf{0.111} & \textbf{0.283} & 0.212 & \textbf{0.378} & 0.325 & 0.338 \\
        \bottomrule
    \end{tabular}
    }
\end{table*}

\begin{table*}[t]
    \caption{Ablation study of Stage 2 heterogeneous evidence scheduling on the CheXpert Plus dataset. All variants are built upon the full Stage 1 setting. Variant \#1 adopts an undifferentiated rigid coupling paradigm. Variants \#2 and \#3 evaluate the individual external evidence branches ($\boldsymbol{E}_{glob}$ and $\boldsymbol{E}_{loc}$, respectively). Variants \#4 and \#5 progressively validate the micro intra-group routing and macro cross-source routing mechanisms.}
    \label{tab:ablation_stage1_stage2_micro}
    \centering
    \setlength{\tabcolsep}{2.3pt}
    \resizebox{\textwidth}{!}{
    \begin{tabular}{lcccccccccccccc}
        \toprule
        \multirow{2}{*}{\raisebox{-0.7ex}{Variant}} & \multirow{2}{*}{\raisebox{-0.7ex}{$\boldsymbol{E}_{loc}$}} & \multirow{2}{*}{\raisebox{-1.1ex}{$\boldsymbol{E}_{glob}$}} & \multirow{2}{*}{\raisebox{-1.8ex}{\makecell{Rigid\\Coupling}}} & \multirow{2}{*}{\raisebox{-1.8ex}{\makecell{Micro\\Routing}}} & \multirow{2}{*}{\raisebox{-1.9ex}{\makecell{Macro\\Routing}}} & \multicolumn{6}{c}{NLG Metrics} & \multicolumn{3}{c}{CE Metrics} \\
        \cmidrule(lr){7-12} \cmidrule(lr){13-15}
        & & & & & & BLEU-1 & BLEU-2 & BLEU-3 & BLEU-4 & ROUGE-L & CIDEr & Precision & Recall & F1 \\
        \midrule
        \#1 & \checkmark & \checkmark & \checkmark &  &  & 0.379 & 0.237 & 0.158 & 0.109 & 0.280 & 0.213 & 0.381 & 0.300 & 0.316 \\
        \#2 &  & \checkmark &  & \checkmark & \checkmark & 0.383 & 0.237 & 0.157 & 0.107 & 0.277 & 0.218 & 0.379 & 0.316 & 0.336 \\
        \#3 & \checkmark &  &  & \checkmark & \checkmark & 0.381 & 0.235 & 0.156 & 0.107 & 0.277 & 0.216 & 0.377 & 0.317 & 0.335 \\
        \#4 & \checkmark & \checkmark &  & \checkmark &  & 0.383 & 0.240 & 0.160 & 0.110 & 0.279 & \textbf{0.220} & \textbf{0.389} & 0.300 & 0.320 \\
        \#5 & \checkmark & \checkmark &  & \checkmark & \checkmark & \textbf{0.383} & \textbf{0.240} & \textbf{0.161} & \textbf{0.111} & \textbf{0.283} & 0.212 & 0.378 & \textbf{0.325} & \textbf{0.338} \\
        \bottomrule
    \end{tabular}
    }
\end{table*}

\begin{figure*}[pos=tb]
    \centering
    \includegraphics[width=\linewidth]{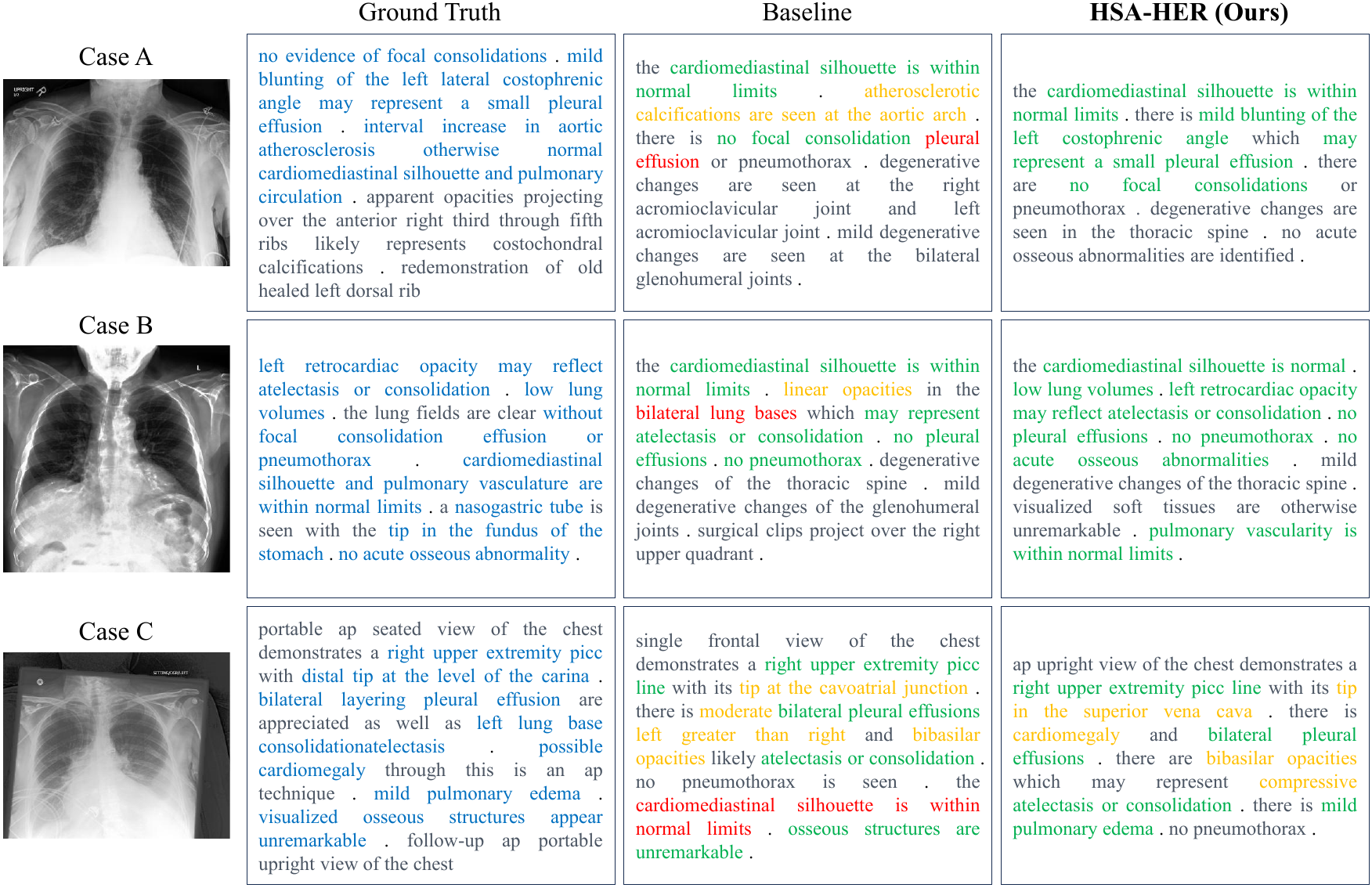}
    \caption{Qualitative comparison on the CheXpert Plus test set. Blue indicates the core evidence in the ground truth. Green represents accurate generations that are semantically consistent. Orange denotes partial matches with deviations in location, severity, certainty, or granularity. Red highlights contradictory statements, such as erroneous negations or unsupported expansions. Dark gray indicates non-core, templated, or unverifiable content.}
    \label{FIG:4}
\end{figure*}

\subsection{Ablation study}

To validate the effectiveness of each core component in the proposed HSA-HER framework, we conduct extensive ablation studies on the CheXpert Plus dataset.

\textbf{Effectiveness of the Explicit Homogeneous Distribution Constraint.}
As shown in Table \ref{tab:ablation_stage1_stage2}, while keeping the Stage 2 modules fixed, the sequential addition of conventional image-text contrastive learning \(\mathcal{L}_{ITC}\) (Variant \#3) and the disease classification constraint \(\mathcal{L}_{CLS}\) (Variant \#4) improves the overall model performance. Specifically, Variant \#4 achieves a Recall of 0.332 and an F1 of 0.345, indicating that these fundamental constraints provide effective cross-modal semantic alignment and disease discrimination capabilities. More importantly, introducing the explicit homogeneous distribution constraint $\mathcal{L}_{Homo}$ to form the complete model (Variant \#7) yields the highest diagnostic Precision (0.378) and optimal results across most NLG metrics (e.g., a BLEU-4 of 0.111 and a ROUGE-L of 0.283). However, the Recall and F1 scores experience a decline. This result demonstrates that the homogeneous distribution constraint does not primarily improve performance by expanding the recall scope for abnormal findings. Instead, it minimizes the distribution shift between the underlying visual and textual representations, thereby enhancing the semantic consistency and discriminative precision of disease-related visual features. The resulting purified visual features establish precise disease semantic anchors, providing reliable semantic guidance for the targeted reconstruction and adaptive weight allocation of heterogeneous clinical evidence in the second stage.

\textbf{Effectiveness of the Hierarchical Expert Routing Mechanism.}
Table \ref{tab:ablation_stage1_stage2} further evaluates the performance gains of the modules in the second stage. Using Variant \#5, which solely contains the complete Stage 1 configuration, as the baseline, the introduction of explicit diagnostic prompts $\boldsymbol{P}_{cls}$ (Variant \#6) increases the Precision from 0.357 to 0.378, indicating that these prompts can provide a certain disease discriminative prior for subsequent report generation. Building upon this, Variant \#7, which fully integrates the hierarchical expert routing, improves the NLG quality while maintaining stable comprehensive clinical efficacy (achieving an F1 of 0.338). This enables sequence matching metrics, such as BLEU-1 to BLEU-4 and ROUGE-L, to reach the highest levels across all variants. These results demonstrate that the hierarchical expert routing enables more effective structured scheduling of heterogeneous clinical evidence, improving the overall quality of the generated reports while maintaining clinical diagnostic performance.

\textbf{Analysis of the Heterogeneous Evidence Scheduling Mechanism.}
To evaluate the heterogeneous evidence scheduling strategy within the hierarchical expert routing, we conducted a targeted ablation study in Table \ref{tab:ablation_stage1_stage2_micro}. Variant \#1 adopts an undifferentiated rigid coupling paradigm to process multi-source features, which causes the Recall to drop to 0.300 and F1 to reach the lowest value of 0.316. This result indicates that without targeted reconstruction and adaptive weight allocation, the undifferentiated rigid coupling strategy fails to utilize the disease semantic anchors established in the first stage. Instead, it dilutes these weak visual abnormality cues during multi-source feature concatenation, thereby weakening the recall capability of the model for abnormal findings. Subsequently, Variants \#2 and \#3 retain only $\boldsymbol{E}_{glob}$ or $\boldsymbol{E}_{loc}$ from the heterogeneous evidence to participate in the routing, increasing their F1 scores to 0.336 and 0.335, respectively. This suggests that even with partial heterogeneous evidence, the routing mechanism can mitigate feature dilution, while maintaining superior comprehensive clinical efficacy compared with the undifferentiated rigid coupling paradigm. Variant \#4 solely employs the micro intra-group routing, achieving the highest Precision and CIDEr (0.389 and 0.220, respectively). This occurs because the micro routing enhances the representation quality of the heterogeneous evidence in each branch through feature reconstruction, yielding excellent performance in diagnostic precision and text consensus. However, due to the lack of dynamic weight balancing at the macro cross-source level, Recall and F1 of the model are only 0.300 and 0.320. Finally, Variant \#5 completely discards the undifferentiated rigid coupling paradigm and introduces the macro cross-source routing for adaptive weighting in coordination with the micro routing. This complete mechanism increases the Recall to 0.325, achieving the overall optimum in comprehensive clinical diagnostic efficacy (an F1 of 0.338).

\subsection{Qualitative Results}
To intuitively evaluate the capability of the models in delineating complex imaging features, we conducted a qualitative comparison between the Baseline and HSA-HER on the CheXpert Plus test set, as shown in Figure ~\ref{FIG:4}. In Case A, faced with the subtle clue of ``mild blunting of the left lateral costophrenic angle'', the Baseline is dominated by high-frequency normal templates and yields an erroneous negative statement. Benefiting from homogeneous semantic alignment, HSA-HER successfully purifies and retains the semantic anchors representing subtle lesions, along with their diagnostic uncertainty. In Case B, the Baseline incorrectly generalizes the opacities to ``bilateral lung bases'', resulting in an unsupported anatomical expansion. Guided by micro intra-group routing, HSA-HER accurately binds the pathological attributes with the ``left retrocardiac'' region, thereby reducing spatial localization hallucinations. In the complex multi-finding scenario of Case C, the Baseline completely misses ``mild pulmonary edema'' and generates factual contradictions, whereas HSA-HER captures more comprehensive clinical findings. Although HSA-HER still exhibits minor deviations in descriptive granularity or certainty under highly complex pathological superpositions, such as outputting the conservative ``possible cardiomegaly'' as a definite diagnosis, its macro cross-source routing mechanism effectively coordinates the adaptive fusion of heterogeneous features. This closely approximates the actual diagnostic logic of clinicians while avoiding severe missed diagnoses. Overall, HSA-HER effectively prevents subtle visual clues from being diluted by text priors and generation inertia, enabling the generated reports to depict complex imaging details more accurately.

\section{Discussion}
This study shows that, compared with methods such as WCL \citep{yan2021weakly} and CAMANet \citep{wang2024camanet}, which mainly improved models through cross-modal alignment, and AM-MRG \citep{wang2026activating} and DADS \citep{wu2026disease}, which expanded diagnostic context through memory or retrieval mechanisms, performance gains in RRG do not depend solely on attention fusion, instance-level contrastive learning, or the introduction of clinical priors. More importantly, they depend on whether precise disease semantic anchors can be established and whether heterogeneous clinical evidence can be effectively scheduled under their guidance. Meanwhile, HSA-HER still shows limitations in Type a, c, d, and e errors under the GREEN evaluation, indicating that improved coverage of key clinical findings is not accompanied by simultaneous improvements in false-positive control, fine-grained localization, severity assessment, and comparative descriptions. Specifically, the current framework is more effective in controlling certain clinically significant errors, while limitations remain in the precise control of diagnostic attributes such as location, severity, and comparative relations. In addition, HSA-HER has mainly been evaluated on two-dimensional chest radiographs, and its fine-grained spatial reasoning for extremely rare lesions requires further investigation. Future work could extend the framework to three-dimensional medical images, such as CT and MRI, as well as longitudinal scans, to further evaluate its applicability to more complex medical imaging scenarios and model the dynamic evolution of diseases.

\section{Conclusion}

In this paper, we propose and evaluate a two-stage homogeneous semantic alignment and hierarchical expert routing framework for RRG. By imposing an explicit homogeneous distribution constraint in the latent space to extract purified visual features as disease semantic anchors, and further using these anchors to guide a hierarchical expert routing mechanism for the targeted reconstruction and adaptive weighting of heterogeneous clinical evidence, our framework effectively overcomes the limitation of traditional RRG paradigms in which weak visual abnormality cues are easily diluted by massive text priors and generation inertia. Experimental results demonstrate that our model achieves excellent performance across three mainstream benchmark datasets. This confirms the effectiveness of the proposed framework in mitigating report templating and accurately depicting complex imaging details and key diagnostic information.

\section*{CRediT authorship contribution statement}

\textbf{Erjian Zhang}: Conceptualization, Software, Validation, Formal analysis, Investigation, Data curation, Writing -- original draft, Writing -- review \& editing, Visualization.
\textbf{Jiayuan Ma}: Conceptualization, Methodology, Software, Validation, Investigation, Writing -- review \& editing.
\textbf{Liejun Wang}: Resources, Writing -- review \& editing, Supervision, Project administration, Funding acquisition.
\textbf{Yikemaiti Sataer}: Writing -- review \& editing, Funding acquisition.
\textbf{Xiaoming Tao}: Writing -- review \& editing, Supervision, Project administration.
\textbf{Zhiqing Guo}: Resources, Writing -- review \& editing, Supervision, Project administration, Funding acquisition.

\section*{Declaration of competing interest}
The authors declare that they have no known competing financial interests or personal relationships that could have appeared to influence the work reported in this paper.

\section*{Declaration of generative AI use}
During the preparation of this work, the authors used ChatGPT (OpenAI) and Gemini (Google) to assist with language polishing and the identification of potentially relevant literature. All identified literature was verified against the original sources, and all AI-assisted outputs were reviewed and edited by the authors as needed. All scientific decisions, experimental results, analyses, and conclusions were independently made and validated by the authors. The authors take full responsibility for the final content of this work.

\section*{Data availability}
All datasets used in this study are publicly accessible, and their sources are described and cited in the manuscript. Part of the source code has already been made publicly available at \url{https://github.com/yyYuyao/HSA-HER}, and the complete source code will be made publicly available upon publication.

\section*{Acknowledgements}
This work was supported in part by the Xinjiang Talent Development Fund under Grant XJRC-2025-KJ-YJ-CXPT-073, in part by the National Natural Science Foundation of China under Grant 62472368, and Grant 62462060.

\bibliographystyle{cas-model2-names}

\bibliography{cas-refs}


\end{document}